\documentclass[11pt, letterpaper, logo, onecolumn, copyright, numbering]{acc}

\usepackage[utf8]{inputenc} 
\usepackage[T1]{fontenc}    
\usepackage{hyperref}       
\usepackage{url}            
\usepackage{booktabs}       
\usepackage{amsfonts}       
\usepackage{nicefrac}       
\usepackage{microtype}      
\usepackage{xcolor}         

\usepackage{graphicx}
\usepackage{caption}
\usepackage{subcaption}
\usepackage{tabularx}
\usepackage{amsmath}
\usepackage{comment}
\usepackage{multirow}
\usepackage{colortbl}
\usepackage{array}
\usepackage{bm}

\newcolumntype{P}[1]{>{\centering\arraybackslash}p{#1}}

\usepackage{natbib}

\usepackage{amsthm}

\usepackage{times}

\definecolor{darkblue}{RGB}{32, 64, 129}
\definecolor{darkgreen}{RGB}{0, 110, 85}
\definecolor{darkred}{RGB}{153, 0, 0}
\definecolor{graytext}{gray}{0.45}

\definecolor{evaluatorcolor}{RGB}{255,230,230}    %
\definecolor{evaluatorframe}{RGB}{180,50,50}      %
\definecolor{taskgencolor}{RGB}{230,255,230}      %
\definecolor{taskgenframe}{RGB}{50,180,50}        %
\definecolor{executioncolor}{RGB}{230,230,255}    %
\definecolor{executionframe}{RGB}{50,50,180}      %
\definecolor{lightgray}{RGB}{240,240,240}
\definecolor{darkgray}{RGB}{80,80,80}

\usepackage{pifont}
\usepackage{longtable}

\usepackage{makecell}

\usepackage{amsmath}
\usepackage{pifont}

\usepackage{booktabs}
\usepackage{multirow}
\usepackage{graphicx}
\usepackage{float}
\usepackage{caption}
\usepackage{subcaption}
\usepackage{xspace}

\usepackage[utf8]{inputenc}
\usepackage[T1]{fontenc}
\usepackage{xcolor}
\usepackage[most]{tcolorbox}
\usepackage{enumitem}
\usepackage{listings}

\definecolor{darkgray}{rgb}{0.3, 0.3, 0.3}
\definecolor{lightgray}{rgb}{0.95, 0.95, 0.95}
\definecolor{codegray}{rgb}{0.98, 0.98, 0.98}

\newtcolorbox{promptbox}[1]{
    colback=lightgray,
    colframe=darkgray,
    colbacktitle=darkgray,
    coltitle=white,
    boxrule=2pt,
    arc=0mm,
    left=10pt,
    right=10pt,
    top=10pt,
    bottom=10pt,
    fonttitle=\bfseries\large,
    title={#1},
    attach boxed title to top left={yshift=-2mm}
}

\usepackage{wrapfig}

\usepackage{algorithmicx}
\usepackage[noend]{algpseudocode}

\definecolor{deepred}{rgb}{0.631,0.102,0.102}
\definecolor{skyblue}{HTML}{126da2}

\definecolor{accpurple}{HTML}{A100FF}
\hypersetup{
     colorlinks = true,
     breaklinks = true,
     linkcolor = accpurple,
     urlcolor = accpurple,
     citecolor = accpurple
     }

\definecolor{orange}{rgb}{1,0.5,0}

\algnewcommand{\LineComment}[1]{\State \(\triangleright\) #1}

\usepackage[ruled,vlined,linesnumbered,noend]{algorithm2e} 

\title{ProRefine: Inference-Time Prompt Refinement with Textual Feedback}

\author[1,2]{Deepak Pandita}
  \author[1]{Tharindu Cyril Weerasooriya}
  \author[1]{Ankit Parag Shah}
  \author[1,3]{Isabelle Diana May-Xin Ng}
  \author[2]{Christopher M. Homan}
  \author[1]{Wei Wei}
  \affil[1]{Center for Advanced
   AI, Accenture}
  \affil[2]{Rochester Institute of Technology}
  \affil[3]{UC Berkeley}

  \date{\today}
  \correspondingauthor{Deepak Pandita (\href{mailto:deepak@mail.rit.edu}{deepak@mail.rit.edu}). Tharindu Cyril Weerasooriya (\href{mailto:t.weerasooriya@accenture.com}{t.weerasooriya@accenture.com}). Work done during internship at Accenture. Work to appear at the Workshop on Efficient Reasoning at NeurIPS 2025. }

\begin{document}
\begin{abstract}
Agentic workflows, where multiple AI agents collaborate to accomplish complex tasks like reasoning or planning, play a substantial role in many cutting-edge commercial applications, and continue to fascinate researchers across fields for their potential to accomplish expensive, complex tasks that, until recently, only humans have been trusted to do. These workflows depend critically on the prompts used to provide the roles models play in such workflows. Poorly designed prompts that fail even slightly to guide individual agents can lead to sub-optimal performance that may snowball within a system of agents, limiting their reliability and scalability. To address this important problem of inference-time prompt optimization, we introduce ProRefine, an innovative inference-time optimization method that uses an agentic loop of LLMs to generate and apply textual feedback. ProRefine dynamically refines prompts for multi-step reasoning tasks without additional training or ground truth labels. Evaluated on five benchmark mathematical reasoning datasets, ProRefine significantly surpasses zero-shot Chain-of-Thought baselines by 3 to 37 percentage points. This approach not only boosts accuracy but also allows smaller models to approach the performance of their larger counterparts. This highlights its potential for building more cost-effective and powerful hybrid AI systems, thereby democratizing access to high-performing AI.
\end{abstract}

\maketitle





\section{Introduction}
\label{sec_intro}

The advancement of Large Language Models (LLMs) is intrinsically linked to their alignment with human values and preferences \citep{feng-etal-2024-llama}. While Reinforcement Learning from Human Feedback (RLHF) has been the cornerstone of this effort \citep{NIPS2017_christiano}, recent research has pivoted towards using LLMs themselves as scalable proxies for human judgment, serving as evaluators, critics, and sources of feedback \cite{NEURIPS2023_zheng, pryzant-etal-2023-automatic, saunders_self-critiquing_2022}. This has given rise to sophisticated agentic frameworks that can detect errors, critique outputs, and iteratively refine them, particularly for tasks demanding factual correctness \cite{akyurek-etal-2023-rl4f, NEURIPS2023_91edff07}. Methods like TextGrad have even demonstrated how textual feedback can ``differentiate'' through complex systems to optimize performance \cite{yuksekgonul2024textgrad}.

\begin{figure}[ht]
    \centering
    \includegraphics[width=0.65\linewidth]{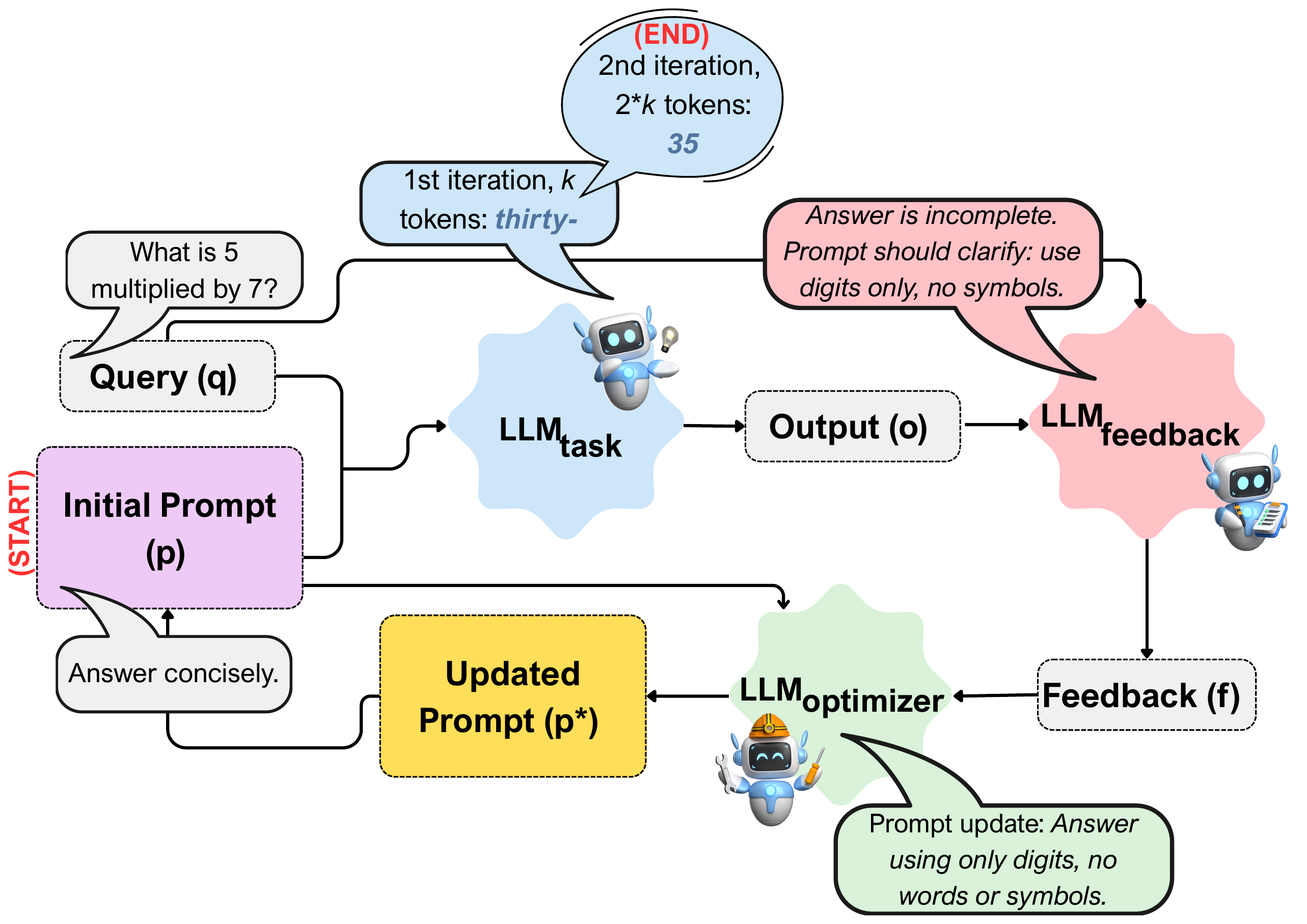}
    \caption{Overview of ProRefine system, illustrating the iterative process of prompt optimization using feedback from LLMs. In each iteration, $LLM_{task}$ extends its output by an additional $k$ tokens, enabling step-by-step feedback to progressively refine the prompt with $LLM_{optimizer}$.}
    \label{fig:system_diagram}
\end{figure}

Our work focuses on optimizing the \textit{prompt}, a key element in chain-of-thought (CoT) \citep{NEURIPS2022_wei} based LLM reasoning. Although prior work has explored prompt optimization \cite{deng-etal-2022-rlprompt, dong-etal-2024-pace, shin-etal-2020-autoprompt, yang2024largelanguagemodelsoptimizers}, they all often focus on either \emph{offline fine-tuning}, which requires extensive training data, or universal application of \emph{largest, most capable models to every task}. This presents a practical dilemma in many real-world scenarios. Continuously fine-tuning is not always feasible, and relying exclusively on state-of-the-art models is often computationally prohibitive. A different approach is needed for scenarios that require \emph{dynamic, on-the-fly repair} for specific and difficult queries where a standard prompt fails. This is particularly true in \emph{resource-aware deployments}, where a smaller model may suffice for most tasks but requires enhancement for a small subset of critical queries. \emph{\textbf{The goal, therefore, shifts from finding a single, universally optimal prompt to performing targeted, inference-time intervention.}}

To address this need, we introduce \textbf{ProRefine} (Inference-time \textbf{Pro}mpt \textbf{Refine}ment with Textual Feedback), which builds upon CoT by adaptively improving prompts using feedback ($LLM_{feedback}$) and an optimizer ($LLM_{optimizer}$) to refine prompts for the task-performing LLM ($LLM_{task}$). This workflow (Figure \ref{fig:system_diagram}), motivated by the teacher-student framework \cite{torrey2013teaching} where a teacher agent guides a student agent to perform a task by providing feedback at intermediate steps, but implemented via LLM interactions without pre-training, represents a novel approach to adaptive agentic reasoning. We explore policy optimization for aligning compound AI systems, drawing inspiration from TextGrad and policy gradient algorithms, such as PPO.




This hybrid-model paradigm makes a method like ProRefine a practical solution. It is designed for resource-constrained environments where deploying the largest models for every query isn't feasible, but temporary access to a capable feedback LLM (perhaps via a separate API call) is possible for critical tasks. In such cases, the refinement process is triggered as an on-demand ``expert intervention.'' ProRefine is task-agnostic and requires no additional training or ground-truth labels. It is an inference-time optimization method that relies on the availability of test-time compute and the ability of LLMs to provide and act upon feedback for optimization.

The ability to break complex tasks into smaller steps and dynamically improve prompts offers a crucial advantage in multi-step agentic workflows where errors can compound.
As illustrated in Figure \ref{fig:example} in the Appendix,
This method is also suitable for black-box LLMs where only API access is available. ProRefine could prove to be crucial in situations demanding greater interpretability, where feedback steps (outputs of $LLM_{feedback}$) offer insights into the reasoning correction process and applications requiring dynamic adaptation without retraining/fine-tuning cycles. To demonstrate its effectiveness, we evaluate ProRefine across five benchmark mathematical reasoning datasets, showing it offers a robust alternative to solely scaling up the base model for all queries.

\paragraph{Key Contributions:}
\begin{itemize}
    \item We propose a novel method - ProRefine - for prompt optimization at \emph{inference-time} using textual feedback.
    \item We evaluated ProRefine on five datasets: object counting, word sorting, grade-school math problem solving, math word problems, and algebraic word problems, and compared our method against CoT and TextGrad.
    \item We evaluate the importance of using a verifier at inference time.
\end{itemize}


\section{ProRefine}
\label{sec_method}

\paragraph{ProRefine} is an inference-time prompt optimization algorithm that optimizes prompts by using textual feedback.
ProRefine involves interactions between three LLMs:

\noindent\bm{$LLM_{task}$}: Executes the task based on the current prompt, generating the initial and subsequent outputs.

\noindent\bm{$LLM_{feedback}$}: A model that critiques the $LLM_{task}$'s output, providing detailed feedback on improvements. This model should be capable of providing insightful and accurate critiques \cite{bai2022constitutional, saunders_self-critiquing_2022}.

\noindent\bm{$LLM_{optimizer}$}: Interprets the feedback and refines the prompt, aiming for coherent and task-focused improvements. This LLM is crucial for ensuring the prompt evolves effectively.

\paragraph{ProRefine} (Algorithm \ref{alg:prorefine}) works as follows\footnote{The code is available at \url{https://github.com/deepakpandita57/ProRefine_public}}:

\begin{algorithm}
\SetKwInput{KwInput}{Input} 
\SetKwInput{KwOutput}{Output} 
\SetKw{Break}{break}
\caption{ProRefine}
\label{alg:prorefine}
\KwInput{Query: $q$, Initial prompt: $p$, tokens\_per\_step: $k$, max\_steps: $n$, LLMs: $LLM_{task}$, $LLM_{feedback}$, $LLM_{optimizer}$}
\KwOutput{Optimized prompt: $p^*$}
$p^*=p$\\
\For{{$i=1$ \KwTo $n$}}{
$o_i = LLM_{task}(p^*, q)$ \tcp{Generate $i*k$ tokens}
$f_i = LLM_{feedback}(q, o_i)$ \tcp{Get textual feedback}
$p* = LLM_{optimizer}(p^*, f_i)$ \tcp{Optimize the prompt}
\If{$EOS\_token$ in $o_i$}{\Break}
}
\Return $p^*$ \tcp{Return final optimized prompt}
\end{algorithm}

\paragraph{Initialization:}

Start with an initial prompt $p$ for the task, a query $q$, and parameters defining the generation and optimization process ($k$ tokens per step, $n$ maximum steps).

\paragraph{Generation and Feedback Loop:}
\begin{itemize}
    \item \textbf{Generation:} Use $LLM_{task}$ to generate an output based on the current prompt $p^*$ and query $q$. This step is limited to $i$ $*$ $k$ tokens to control the granularity of the feedback. In each iteration, $LLM_{task}$ produces $k$ more tokens, attempting to refine prior output while progressively continuing its response to the query.
    \item \textbf{Feedback:} $LLM_{feedback}$ evaluates the generated output $o_i$ against the query $q$ to provide textual feedback $f_i$. This feedback encapsulates how the output could be improved, focusing on aspects such as accuracy, relevance, or coherence.
    \item \textbf{Optimization:} $LLM_{optimizer}$ uses the feedback $f_i$ to refine the prompt $p^*$. This step involves modifying the prompt to better align with the task requirements or to correct identified deficiencies in previous generations.
\end{itemize}

\paragraph{Termination:}
The process iterates until either the maximum number of steps $n$ is reached or an end-of-sequence (EOS) token is detected in the output, indicating the completion of the task.

The granularity and duration of the optimization process are governed by two parameters: $k$, the number of tokens per step, and $n$, the maximum number of steps. These parameters can be adjusted according to the task's complexity and the desired output quality. For example, rather than generating feedback every $k$ tokens, we might instead choose to provide feedback after each sentence or paragraph, particularly in tasks such as machine translation or text summarization, where larger semantic units may be more meaningful.

\paragraph{Unifying Verifier and Feedback:}
At inference time, verifiers play a crucial role in judging model outputs \cite{cobbe2021training, lightman_lets_2024, snell_scaling_2024}.
For simplicity in this study, we do not train a bespoke verifier; rather, we employ the \textit{Llama3.1-70B-instruct} model to function as both the feedback mechanism ($LLM_{feedback}$) and the verifier. We manage these roles through separate API calls, each with a role-defining prompt. A smaller model, specifically fine-tuned for these tasks, could also be used. The verifier's function is to evaluate the initial output generated by $LLM_{task}$ for each query. If the verifier assesses the output to be incorrect, the refinement process is triggered; otherwise, the output is used as is. This also saves computation on answers that are already correct.

To quantify the verifier's impact, we analyze three distinct scenarios: \emph{ProRefine (verifier)}, our standard approach which employs $LLM_{feedback}$ to guide refinement; \emph{ProRefine (no verifier)}, wherein the refinement process operates without verifier input; and \emph{ProRefine (optimal verifier)}, guided by a perfect verifier (simulated using ground-truth labels). This optimal condition reveals the upper bound of the refinement loop's potential. Consequently, the performance difference between \emph{ProRefine (verifier)} and \emph{ProRefine (optimal verifier)} underscores the significance of verifier accuracy. It is important to note that ProRefine's methodology does not inherently rely on labels or optimal verification, despite their use in this specific evaluation.



\section{Experiments and Evaluation}
\label{sec_exp_neurips}

\begin{table*}[!ht]
\centering
\tiny
\begin{tabular}{l|l|c|c|c}
\textbf{Dataset} & \textbf{Method} & \textbf{Llama-3.2 1B-it} & \textbf{Llama-3.2 3B-it} & \textbf{Llama-3.1 8B-it}\\
\hline

\multirow{5}{*}{Object Counting}
& CoT                 & 0.48 [0.382, 0.578] & 0.65 [0.556, 0.744] & 0.73 [0.643, 0.817] \\
& TextGrad           & \textbf{0.62} [0.524, 0.716] & 0.73 [0.643, 0.817] & 0.86 [0.792, 0.928] \\
& ProRefine (no verifier)       & 0.51 [0.412, 0.608] & \textbf{0.75} [0.665, 0.835] & 0.77 [0.687, 0.853] \\
& \cellcolor{gray!20}ProRefine (verifier)    & \cellcolor{gray!20}0.6 [0.503, 0.696] & \cellcolor{gray!20}0.72 [0.632, 0.808] & \cellcolor{gray!20}\textbf{0.89}* [0.839, 0.959] \\
\cline{2-5}
& \textsuperscript{\textdagger}ProRefine (optimal verifier)    & 0.67 [0.577, 0.763] & 0.85* [0.780, 0.920] & 0.94* [0.893, 0.987] \\
\hline

\multirow{5}{*}{Word Sorting}
& CoT               & 0.11 [0.048, 0.172] & 0.10 [0.041, 0.159] & 0.50 [0.401, 0.598] \\
& TextGrad         & \textbf{0.33}* [0.237, 0.423] & \textbf{0.61}* [0.514, 0.706] & 0.69* [0.599, 0.781] \\
& ProRefine (no verifier)     & 0.22 [0.138, 0.302] & 0.47* [0.372, 0.568] & 0.68 [0.595, 0.779] \\
& \cellcolor{gray!20}ProRefine (verifier)  & \cellcolor{gray!20}0.19 [0.113, 0.267] & \cellcolor{gray!20}0.32* [0.228, 0.412] & \cellcolor{gray!20}\textbf{0.71}* [0.621, 0.799] \\
\cline{2-5}
& \textsuperscript{\textdagger}ProRefine (optimal verifier)  & 0.29* [0.192, 0.368] & 0.53* [0.432, 0.628] & 0.86** [0.792, 0.928] \\
\hline

\multirow{5}{*}{GSM8K}
& CoT               & 0.450 [0.423, 0.476] & 0.809 [0.787, 0.829] & 0.819 [0.797, 0.839] \\
& TextGrad         & 0.463 [0.436, 0.489] & 0.801 [0.779, 0.822] & 0.864* [0.845, 0.882] \\
& ProRefine (no verifier)     & 0.636** [0.610, 0.662] & 0.797 [0.774, 0.818] & 0.843 [0.823, 0.863 \\
& \cellcolor{gray!20}ProRefine (verifier)  & \cellcolor{gray!20}\textbf{0.654}** [0.627, 0.678] & \cellcolor{gray!20}\textbf{0.866}** [0.847, 0.883] & \cellcolor{gray!20}\textbf{0.885}* [0.868, 0.902] \\
\cline{2-5}
& \textsuperscript{\textdagger}ProRefine (optimal verifier)  & 0.725** [0.701, 0.749] & 0.904** [0.888, 0.920] & 0.936** [0.922, 0.949] \\
\hline

\multirow{5}{*}{SVAMP}
& CoT               & 0.689 [0.66, 0.718] & 0.869 [0.848, 0.890] & 0.854 [0.832 , 0.876] \\
& TextGrad         & 0.684 [0.655, 0.713] & 0.861 [0.840, 0.882] & 0.84 [0.817, 0.863] \\
& ProRefine (no verifier)     & 0.774** [0.748, 0.800] & 0.878 [0.858, 0.898] & 0.877 [0.857, 0.897] \\
& \cellcolor{gray!20}ProRefine (verifier)  & \cellcolor{gray!20}\textbf{0.808}** [0.784, 0.832] & \cellcolor{gray!20}\textbf{0.896} [0.877, 0.915] & \cellcolor{gray!20}\textbf{0.893}* [0.874, 0.912] \\
\cline{2-5}
& \textsuperscript{\textdagger}ProRefine (optimal verifier)  & 0.861** [0.840, 0.882] & 0.925** [0.909, 0.941] & 0.938** [0.923, 0.953] \\
\hline

\multirow{5}{*}{AQUARAT}
& CoT               & 0.259 [0.202, 0.31] & \textbf{0.563} [0.498, 0.620] & 0.586 [0.522, 0.643] \\
& TextGrad         & \textbf{0.311} [0.250, 0.364] & 0.524 [0.462 , 0.585] & 0.559 [0.494, 0.616] \\
& ProRefine (no verifier)     & 0.205 [0.151, 0.250] & 0.343 [0.284, 0.401] & 0.398 [0.337 , 0.458] \\
& \cellcolor{gray!20}ProRefine (verifier)  & \cellcolor{gray!20}0.268 [0.209, 0.318] & \cellcolor{gray!20}0.551 [0.486 , 0.608] & \cellcolor{gray!20}\textbf{0.606} [0.542, 0.663] \\
\cline{2-5}
& \textsuperscript{\textdagger}ProRefine (optimal verifier)  & 0.354 [0.292, 0.409] & 0.598 [0.538, 0.659] & 0.657 [0.595, 0.712 ] \\

\end{tabular}
\caption{Test Accuracy with 95\% confidence intervals across five benchmark datasets and models. * and ** denote statistically significant improvements over one or two baseline methods, respectively. Results in bold indicate the highest accuracy for a dataset-method combination. \textsuperscript{\textdagger} demonstrates the upper bound potential of the optimization loop and the impact of verifier quality. \textit{Llama3.1-70B-instruct} is employed for feedback generation, prompt optimization, and evaluation.}
\label{tab:results}
\end{table*}


\subsection{Data}

We evaluate ProRefine on five reasoning tasks, each of which involves multi-step reasoning, making them suitable for evaluating prompt optimization in agentic workflows. We utilize object counting and word sorting from the BIG-Bench Hard benchmark \cite{srivastava2023beyond}, grade-school math problem-solving from GSM8K \cite{cobbe2021training}, math word problems from SVAMP \cite{patel-etal-2021-nlp}, and algebraic word problems from AQUARAT \cite{ling-etal-2017-program}. See Appendix \ref{sec_exp} for details about data splits.

\subsection{Experimental Setup}
\label{sec_setup_neurips}

We experiment with three models - \textit{Llama3.2-1B-instruct}, \textit{Llama3.2-3B-instruct}, and \textit{Llama3.1-8B-instruct} \cite{llama3_2_model_card} for $LLM_{task}$.
The prompts are optimized using ProRefine, with \textit{Llama3.1-70B-instruct} used for feedback generation, prompt optimization, and evaluation. We select the values of hyperparameters $k=10$ and $n=25$ to control the granularity of feedback and duration of optimization. Hyperparameters $k$ and $n$ were fixed based on general preliminary exploration and not tuned per task using benchmark training/validation data.

We compare ProRefine against the zero-shot Chain-of-Thought (CoT) baseline and TextGrad \cite{yuksekgonul2024textgrad}, and report test accuracy with 95\% confidence interval.
It is essential to remember that TextGrad is a supervised fine-tuning method that utilizes both the training and validation sets.

\subsection{Results}
\label{sec_res_neurips}

Our results (Table \ref{tab:results}) demonstrate that ProRefine significantly improves $LLM_{task}$ performance over the zero-shot CoT baseline in all but one experiment, and it outperforms TextGrad in 11 out of 15 cases overall. For \textit{Llama3.2-1B-instruct} model, ProRefine can significantly outperform CoT and TextGrad on 2 out of 5 datasets. For \textit{Llama3.2-3B-instruct} model, ProRefine can outperform CoT and TextGrad on 3 out of 5 datasets with one significant result. For \textit{Llama3.1-8B-instruct} model, ProRefine can outperform CoT and TextGrad on all 5 datasets with 4 significant results.

\paragraph{Object Counting:} ProRefine improves performance by $3-16$ percentage points over CoT, with significant gains observed for \textit{Llama3.1-8B-instruct}. It outperforms TextGrad on 2 out of 3 models, yielding a $2-3$ percentage point advantage. However, a performance drop of $2$ points is observed for \textit{Llama3.2-1B-instruct}. 

\paragraph{Word Sorting:} Performance gains over CoT range from $8-37$ percentage points, with significant improvements for \textit{Llama3.2-3B-instruct} and \textit{Llama3.1-8B-instruct}. ProRefine surpasses TextGrad on 1 of 3 models with a $2$-point gain, but performance drops of $11-14$ points are observed for \textit{Llama3.2-1B-instruct} and \textit{Llama3.2-3B-instruct}.

\paragraph{GSM8K:} ProRefine achieves $2.4-20.4$ percentage points improvement over CoT, with significant improvement observed for all the models; however, a slight performance drop ($1.2$) is observed for \textit{Llama3.2-3B-instruct}. It outperforms TextGrad on all models, achieving a $2.1-19.1$ percentage point gain with significant results observed for \textit{Llama3.2-1B-instruct} and \textit{Llama3.2-3B-instruct} models. Minor performance drop of $0.4-2.1$ is observed for \textit{Llama3.2-3B-instruct} and \textit{Llama3.1-8B-instruct}.

\paragraph{SVAMP:} Performance improves by $0.9-11.9$ percentage points over CoT, with significant gains for \textit{Llama3.2-1B-instruct} and \textit{Llama3.1-8B-instruct}. ProRefine outperforms TextGrad across all models, with $1.7-12.4$ percentage point gains and significant results for \textit{Llama3.2-1B-instruct}.

\paragraph{AQUARAT:} Gains over CoT range from $0.9-2$ percentage points, but declines of $5.4-22$ points are also observed. ProRefine exceeds TextGrad on 2 of 3 models, with $2.7-4.7$ percentage point gains, although performance drops of $10.6-18.1$ points are also recorded.

Our results demonstrate that using ProRefine with an optimal verifier significantly improves performance for all tasks, achieving the best results in 13 out of 15 cases, highlighting the critical role of verifier quality. Notably, the number of significant improvements increases with larger model sizes.
We also observe that ProRefine enables smaller models, such as \textit{Llama3.2-3B-instruct} and \textit{Llama3.1-8B-instruct}, to approach the zero-shot performance of larger models like \textit{Llama3.1-8B-instruct} and \textit{Llama3.1-70B-instruct}, respectively.


\section{Related Work}
\label{sec_rel_work}

ProRefine draws inspiration from and contributes to several interconnected research areas. The performance of LLMs is heavily dependent on the quality of the prompts they receive. Early efforts in this domain centered on crafting prompts manually \cite{NEURIPS2022_wei}, a meticulous process of designing effective prompts to elicit desired responses. Recognizing the limitations and scalability challenges of manual methods, research has increasingly focused on automatic prompt optimization with a growing emphasis on agentic workflows that enable dynamic and adaptive reasoning. 

\paragraph{Prompt Generation:}
Some pioneering automatic methods, such as AutoPrompt \cite{shin-etal-2020-autoprompt} and RLPrompt \cite{deng-etal-2022-rlprompt}, employ gradient-based search and reinforcement learning techniques, respectively. AutoPrompt \cite{shin-etal-2020-autoprompt} uses gradient-based search to generate prompts for masked language models. It reformulates tasks as fill-in-the-blank problems, achieving performance comparable to supervised models in tasks like sentiment analysis. However, it requires training data and gradient access, limiting its applicability to black-box models. Other approaches leverage LLMs themselves for prompt generation \cite{mehta2024promptly, pryzant-etal-2023-automatic, yang2024largelanguagemodelsoptimizers, yang-etal-2022-re3, zhou2022large}. Recent works like Promptomatix \cite{murthy2025promptomatixautomaticpromptoptimization} and EvoAgentX \cite{wang2025evoagentxautomatedframeworkevolving} extend this direction by enabling automatic prompt refinement across multiple tasks, workflows, and tools. ProRefine distinguishes itself by operating simply at inference-time, requiring no training data, gradient access, or model retraining, while enabling prompt refinement in dynamically evolving settings.

\paragraph{Self-Refinement:}
There is a substantial and growing body of work exploring the capacity of LLMs to act as judges or evaluators  \cite{bavaresco2024llmsinsteadhumanjudges, chiang-lee-2023-large, li-etal-2024-leveraging-large, liu-etal-2023-g, verga2024replacing, wang-etal-2023-chatgpt, NEURIPS2023_zheng, zhuge2024agentasajudgeevaluateagentsagents}. This capability has been leveraged to assess response quality or provide self-feedback. ProRefine adopts this principle, using LLM-generated textual feedback to improve its own prompting process. Unlike prior uses of LLM evaluation solely for ranking or filtering, ProRefine uses that feedback in a closed-loop for optimization during inference.

The idea of LLM iterative refinement is highly relevant. Self-Refine \cite{NEURIPS2023_91edff07} is a prominent example, where an LLM generates both output and feedback, using the latter for refinement. ARIES \cite{zeng2025evolvingllmsselfrefinementcapability} further enhances refinement via Elo-style agent debate. Other works explore self-critiquing \cite{saunders_self-critiquing_2022} and reinforcement learning for critique generation (RL4F) \cite{akyurek-etal-2023-rl4f}, along with various feedback and refinement mechanisms \cite{dong-etal-2024-pace, khattab2024dspy, qu2024recursive, ranaldi2024self, schick2023peer, wadhwa-etal-2024-learning-refine}, and Monte Carlo-based refinement in math reasoning (MC-NEST) \cite{rabby2025mcnestenhancingmathematicalreasoning}.  While ProRefine shares the self-refinement spirit, it focuses on prompt refinement, suitable for agentic workflows and black-box LLMs, while avoiding reinforcement learning and direct output modification.
 
 \paragraph{Inference-Time Scaling:}
ProRefine belongs to the broader category of inference-time methods  \cite{muennighoff2025s1, snell2024scalingllmtesttimecompute},  that improve LLMs without without weight modification \cite{du2024improving}. Inference-time methods aim to improve the performance of models by utilizing test-time compute resources. TextGrad \cite{yuksekgonul2024textgrad} performs gradient-free inference-time optimization using textual feedback. ProRefine applies a similar idea to intermediate prompt refinement for dynamic reasoning chains. TextGrad relies on supervised fine-tuning, whereas ProRefine operates without training data, offering ease of integration. Other inference-time strategies include RL-of-Thoughts \cite{hao2025rlthoughtsnavigatingllm} and Reward-Is-Enough \cite{song2025rewardenoughllmsincontext}, which apply RL-based signal propagation during inference. AvR (Alignment via Refinement) \cite{zhang2025unlockingrecursivethinkingllms} proposes recursive CoT refinement using long-form reasoning. ProRefine, by contrast, performs step-level feedback on prompts rather than final outputs, and requires no external tools or supervision.

\paragraph{Agentic Workflows:}
ProRefine also fits into a broader trend toward agentic workflows. AFlow \cite{zhang2025aflowautomatingagenticworkflow} automates agentic workflows through prompt-based search over prior structures, while EvoAgentX \cite{wang2025evoagentxautomatedframeworkevolving} evolves agent behaviors and topologies. Meanwhile, Mass \cite{zhou2025multiagentdesignoptimizingagents} and DebFlow \cite{su2025debflowautomatingagentcreation} optimize multi-agent configurations via interleaved search and debate. ProRefine focuses instead on optimizing individual agent prompts within fixed workflows, complementing these methods. Unlike tool-integrated or debate-based systems, ProRefine remains model-agnostic and easy to integrate into any prompt-based agent loop. 


\section{Experiments and Evaluation}
\label{sec_exp}

\subsection{Data}

We evaluate ProRefine on five reasoning tasks, each of which involves multi-step reasoning, making them suitable for evaluating prompt optimization in agentic workflows. We include the original dataset split sizes in (train/validation/test) format: object counting and word sorting from the BIG-Bench Hard benchmark \cite{srivastava2023beyond} (50/100/100), grade-school math problem-solving from GSM8K \cite{cobbe2021training} (200/300/1319), math word problems from SVAMP \cite{patel-etal-2021-nlp} (2516/622/1000), and algebraic word problems from AQUARAT \cite{ling-etal-2017-program} (97467/254/254). We use the same splits and evaluation as \citet{yuksekgonul2024textgrad} for object counting, word sorting, and GSM8K.

\subsection{Experimental Setup}
\label{sec_setup}

\begin{figure*}[htb]
    \centering
    \includegraphics[width=0.99\linewidth]{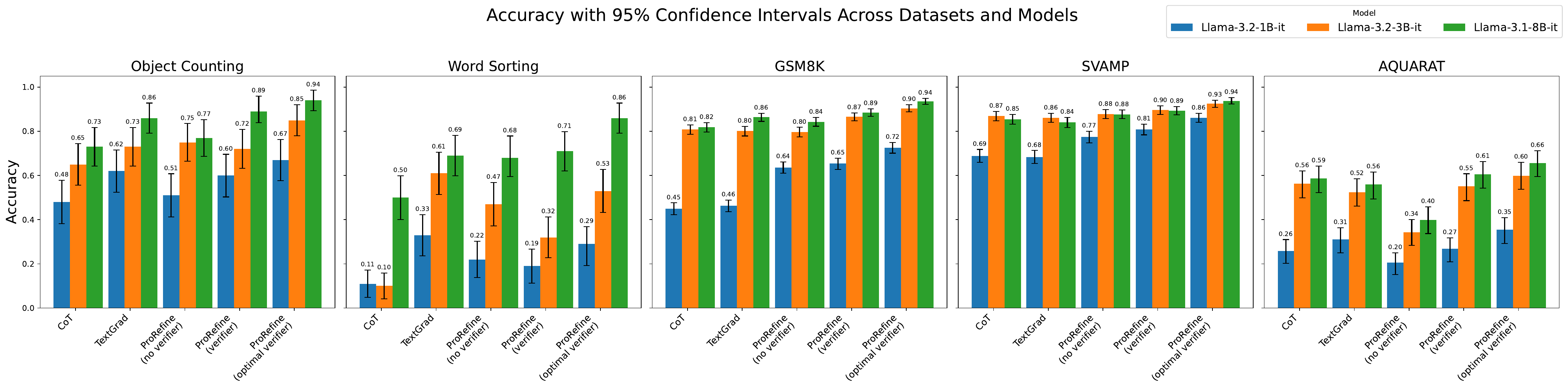}
    \caption{Test Accuracy [with 95\% confidence interval] across different models and datasets. \textit{Llama3.1-70B-instruct} is employed for feedback generation, prompt optimization, and evaluation.}
    \label{fig:results}
\end{figure*}

\begin{figure}[ht]
    \centering
    \includegraphics[width=0.60\linewidth]{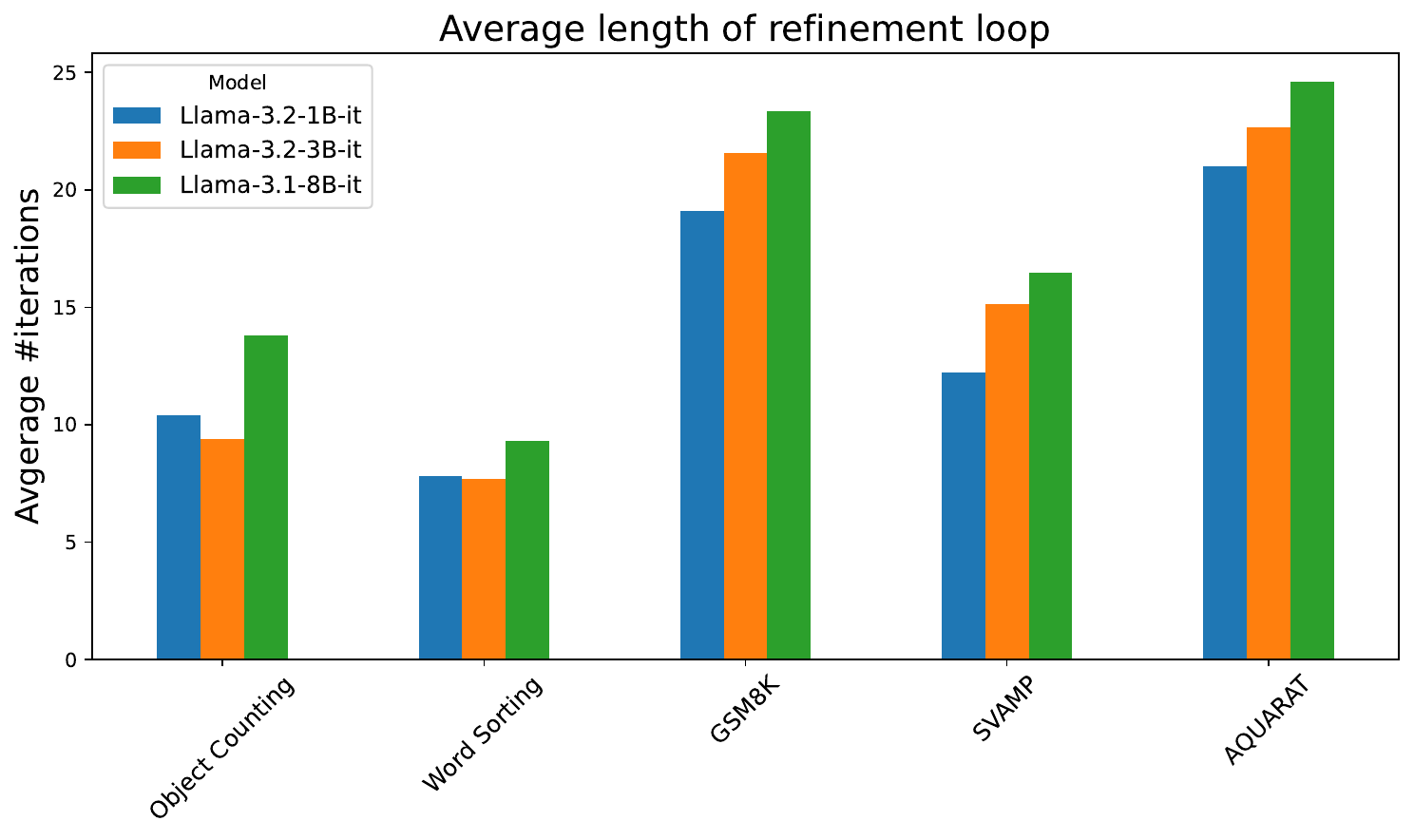}
    \caption{Average number of prompt refinement iterations.}
    \label{fig:ref_stats}
\end{figure}

We experiment with three models - \textit{Llama3.2-1B-instruct}, \textit{Llama3.2-3B-instruct}, and \textit{Llama3.1-8B-instruct} \cite{llama3_2_model_card} for $LLM_{task}$. The prompts are optimized using Algorithm \ref{alg:prorefine}, with \textit{Llama3.1-70B-instruct} used for feedback generation, prompt optimization, and evaluation. We select the values of hyperparameters $k=10$ and $n=25$ to control the granularity of feedback and duration of optimization. Hyperparameters $k$ and $n$ were fixed based on general preliminary exploration and not tuned per task using benchmark training/validation data.

We compare the performance of our method against the zero-shot Chain-of-Thought (CoT) baseline and TextGrad \cite{yuksekgonul2024textgrad}, and report test accuracy with 95\% confidence interval. We choose TextGrad as a baseline because \citet{yuksekgonul2024textgrad} reported performance at par or better than DSPy \cite{khattab2024dspy} for prompt optimization on object counting, word sorting, and GSM8k datasets. It is essential to remember that TextGrad is a supervised fine-tuning method that utilizes both the training and validation sets. For TextGrad, we use a comparative setup consisting of a task model to be fine-tuned and \textit{Llama3.1-70B-instruct} model for feedback generation and backpropagation. The results are shown in Table \ref{tab:results} and Figure \ref{fig:results}.

\begin{figure}[ht!]
    \centering
    \includegraphics[width=0.60\linewidth]{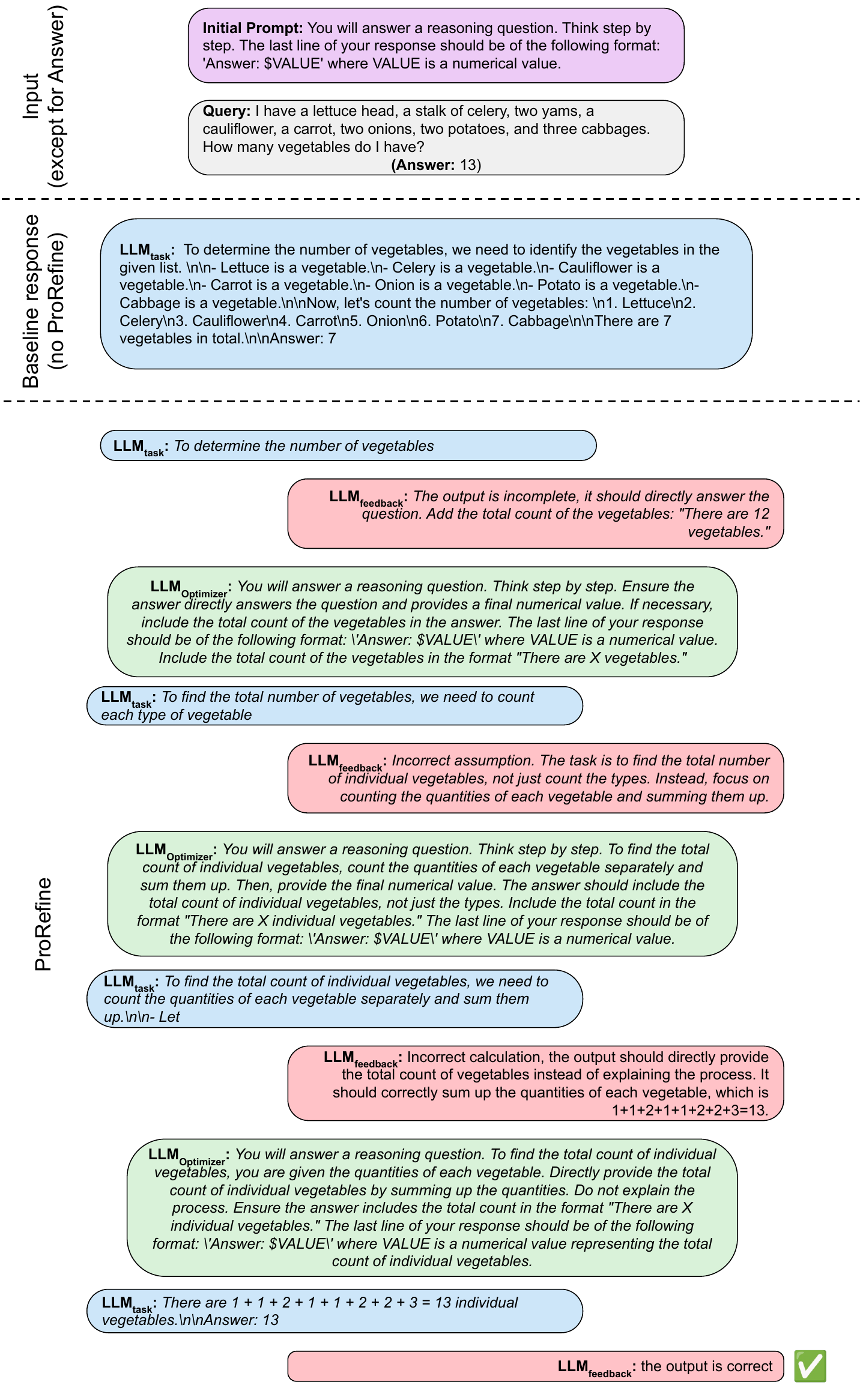}
    \caption{ProRefine example. Given an input query (which in this case has a correct answer: 13) and an initial prompt, the task model ($LLM_{task}$) gives an incorrect answer. ProRefine uses two additional models, $LLM_{feedback}$ and $LLM_{optimizer}$, to iteratively improve the prompt \emph{as the $LLM_{task}$ generates its response}. Refining the prompt during generation allows the feedback model to target local regions of the response, providing finer-grained feedback than waiting for the response to complete. We also provide an additional example illustrating our approach in Figure~\ref{fig:neg_example}.
    \label{fig:example}}
\end{figure}

\section{Discussion}
\label{sec_discussion}

This work investigates the following research questions. 

\paragraph{RQ1} \textit{How effectively can textual feedback enhance the performance of LLMs during inference?} 
\paragraph{RQ2} \textit{To what extent does model size impact the ability of LLMs to utilize textual feedback?}
\paragraph{RQ3} \textit{What is the impact of incorporating a verifier on accuracy at inference time?}

Regarding RQ1, the results demonstrate that ProRefine is a broadly applicable method that utilizes textual feedback to improve LLM performance at inference time. The “performance gap bridging” effect is particularly noteworthy, suggesting that ProRefine may serve as an effective alternative to simply scaling up model size, potentially avoiding costly fine-tuning an advantage in resource-constrained settings.

The largest performance gains are observed on the word sorting task, indicating that tasks requiring more complex reasoning or manipulation of intermediate outputs benefit the most from ProRefine's iterative refinement. The mixed results when using a smaller model for $LLM_{feedback}$ illustrate the importance of ``knowledge asymmetry,'' i.e., that the feedback model should be ``sufficiently capable'' of providing useful critiques.

Regarding RQ2, the results indicate that ProRefine outperforms the baselines on 2 and 3 datasets when using the \textit{Llama3.2-1B-instruct} and \textit{Llama3.2-3B-instruct} models, respectively, and on all 5 datasets when using the \textit{Llama3.1-8B-instruct} model. This suggests that performance improvements scale with model size. These findings imply that larger models are preferable to smaller ones, particularly in agentic workflows that may require test-time scaling and the effective use of textual feedback to solve complex tasks.

\begin{figure}[ht!]
    \centering
    \includegraphics[width=0.60\linewidth]{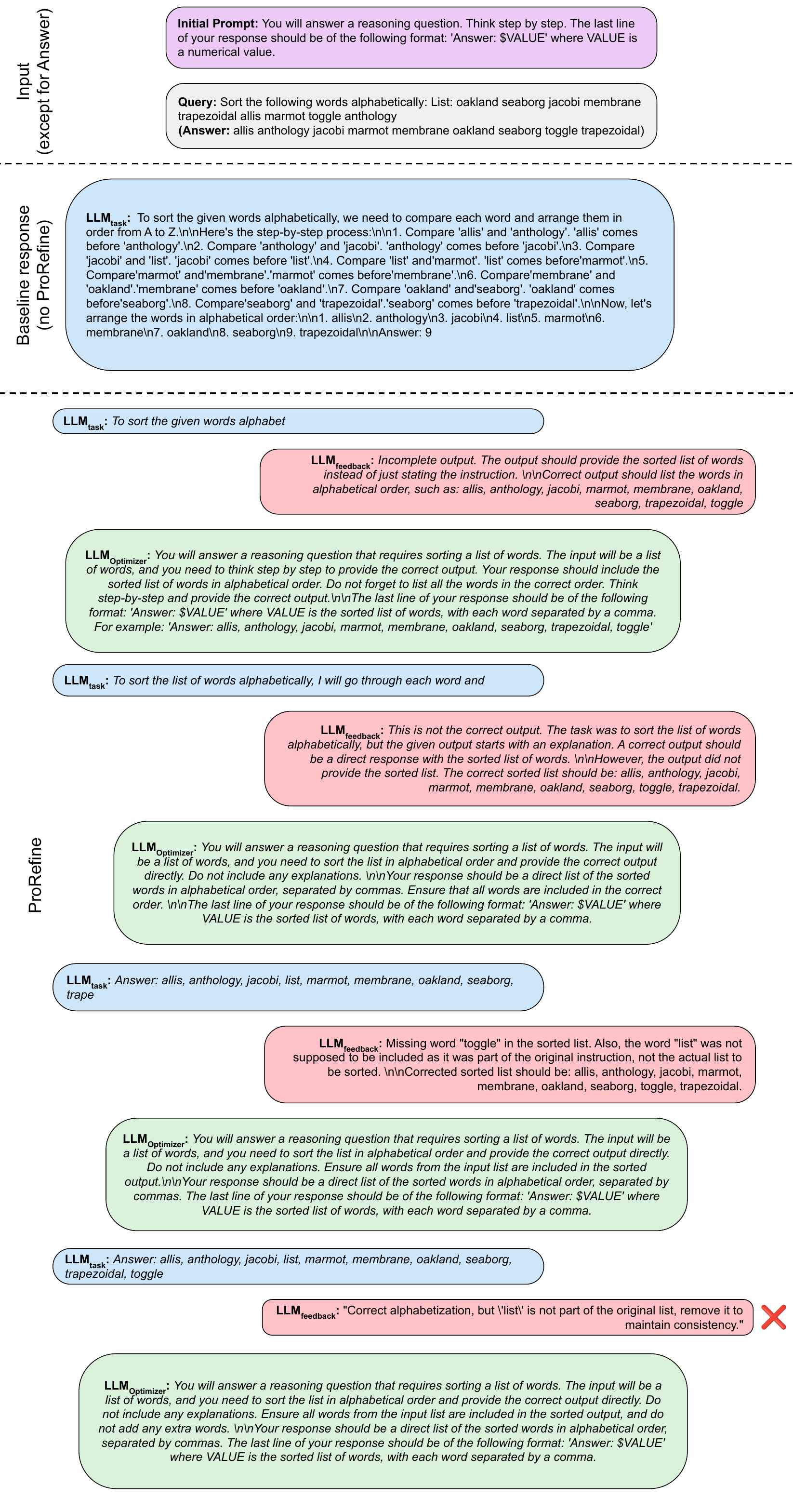}
    \caption{This figure is an instance when $LLM_{optimizer}$ is not aligned with the feedback from $LLM_{feedback}$ and misses important guiding steps. The framework is similar to Figure \ref{fig:example} when $LLM_{optimizer}$ conveys feedback effectively. We've observed a few failed instances following this pattern.}
    \label{fig:neg_example}
\end{figure}

Regarding RQ3, the results highlight that employing a high-quality verifier is crucial for significantly improving task performance at inference time. We observe some cases where ``no verifier'' outperforms the ``verifier'' setting, which indicates the verifier incorrectly accepted a flawed initial answer, thereby preventing the refinement process from correcting the error. This reveals a trade-off: the verifier reduces computational cost on correct answers but risks prematurely halting on incorrect ones. The superior results of the ``optimal verifier'' highlight the critical role of verifier accuracy. Beyond enhancing performance, the verifier also reduces computational cost during inference by guiding the refinement process. Moreover, it opens up promising avenues for future work, where an optimizer could be designed to maximize rewards guided by the verifier’s assessments. ProRefine can offer a degree of interpretability by exposing the outputs from $LLM_{feedback}$, allowing insights into the model's reasoning process. Figures \ref{fig:example} and \ref{fig:neg_example} demonstrate cases where model feedback successfully improves the output and where it fails, respectively. Although evaluated on reasoning and math tasks, ProRefine is general and applicable to other tasks.

A crucial consideration for ProRefine is the trade-off between its accuracy gains and the increased computational cost at inference time. Each refinement step requires additional calls to the $LLM_{feedback}$ and $LLM_{optimizer}$, making any single query more expensive to process than a standard single-pass generation. However, this per-query cost should be evaluated within ProRefine's intended hybrid-model deployment. The strategy is not to run refinement on every query, but to use it as an on-demand intervention precisely when a more efficient base model fails. Therefore, the overall system cost is not a simple sum of expensive refinement processes. Instead, it is a blend of low-cost successes from the base model and high-value, targeted corrections. Moreover, the cost is still considerably lower than full model retraining or fine-tuning. Our results support this approach's practicality: Figure \ref{fig:ref_stats} shows that the average number of refinement iterations is typically low, ensuring the per-incident cost of intervention is contained. This cost-accuracy balance can be further optimized by tuning hyperparameters like feedback granularity ($k$) and maximum iterations ($n$).

\section{Limitations and Future Work}

This work has the following limitations that we acknowledge have potential for future explorations:

\begin{itemize}
    \item \textbf{Computational Cost and Practicality:} While ProRefine is designed for cost-effective hybrid deployments, its iterative process inherently increases inference-time latency and computational cost compared to a single-pass query. The cost-benefit of this trade-off must be carefully evaluated for each specific application, as its viability depends on the base model's failure rate and the relative costs of the LLMs involved.
    \item \textbf{Generalizability:} Our evaluation is currently focused on mathematical and multi-step reasoning tasks. Further research is needed to assess performance across a broader range of reasoning tasks and domains. Our method is also sensitive to hyperparameters and requires manual tuning. Developing more robust, automated, or adaptive methods for setting parameters would enhance the method's usability.
    \item \textbf{Dependence on High-Quality Feedback:} The system's performance is dependent on the quality of the $LLM_{feedback}$. Future work could explore using a specialized ``critic'' model or fine-tuning feedback models to improve diagnostic accuracy. Furthermore, using LLMs for evaluation introduces potential biases and more comprehensive human evaluations and robust methods are need for mitigating evaluator bias.
    \item \textbf{Stability of the Refinement Loop:} The iterative nature of ProRefine lacks a formal convergence guarantee. In some cases, the refinement process can suffer from prompt degradation after many iterations or plateau before reaching an optimal solution. Investigating methods to ensure stable and monotonic improvement is a key area for future research.
\end{itemize}

\section{Conclusion}
\label{sec_conclusion}

We introduced ProRefine, a novel, practical, and \emph{inference-time} prompt optimization method for agentic workflows. ProRefine leverages LLM-generated textual feedback to dynamically refine prompts, leading to significant performance improvements on multi-step reasoning tasks without requiring additional training or ground-truth labels. Our results demonstrate its ability to bridge the performance gap between smaller and larger LLMs, making it a key enabler for more efficient and cost-effective hybrid-model deployments. The \emph{inference-time} nature of ProRefine makes it readily deployable for on-demand reasoning correction, contributing to more adaptable and accessible AI systems. Future work will explore applying this framework to new domains, developing more sophisticated feedback and optimizer agents, and exploring adaptive policies for hyperparameter tuning to further optimize the cost-performance trade-off.


\bibliography{anthology, custom, references}










\end{document}